\documentclass[10pt,twocolumn,letterpaper]{article}

\usepackage{iccv}
\usepackage{times}
\usepackage{epsfig}
\usepackage{graphicx}
\usepackage{amsmath}
\usepackage{amssymb}
\usepackage[table,xcdraw]{xcolor}
\usepackage{multirow}
\usepackage[normalem]{ulem}
\useunder{\uline}{\ul}{}
\usepackage{cite}

\usepackage[accsupp]{axessibility}  

\usepackage[breaklinks=true,bookmarks=false]{hyperref}

\iccvfinalcopy 

\def\iccvPaperID{11914} 

\ificcvfinal\fi

\begin{document}

\title{Fast Inference and Update of Probabilistic Density Estimation \\on Trajectory Prediction}

\author{Takahiro Maeda \qquad Norimichi Ukita\\
Toyota Technological Institute, Japan\\
{\tt\small \{sd21601, ukita\}@toyota-ti.ac.jp}
}

\maketitle
\global\csname @topnum\endcsname 0
\global\csname @botnum\endcsname 0
\ificcvfinal\thispagestyle{empty}\fi

\begin{abstract}
   Safety-critical applications such as autonomous vehicles and social robots require fast computation and accurate probability density estimation on trajectory prediction.
   To address both requirements, this paper presents a new normalizing flow-based trajectory prediction model named FlowChain.
   FlowChain is a stack of conditional continuously-indexed flows (CIFs) that are expressive and allow analytical probability density computation.
   This analytical computation is faster than the generative models that need additional approximations such as kernel density estimation.
   Moreover, FlowChain is more accurate than the Gaussian mixture-based models due to fewer assumptions on the estimated density.
   FlowChain also allows a rapid update of estimated probability densities.
   This update is achieved by adopting the \textit{newest observed position} and reusing the flow transformations and its log-det-jacobians that represent the \textit{motion trend}.
   This update is completed in less than one millisecond because this reuse greatly omits the computational cost.
   Experimental results showed our FlowChain achieved state-of-the-art trajectory prediction accuracy compared to previous methods.
   Furthermore, our FlowChain demonstrated superiority in the accuracy and speed of density estimation.
   Our code is available at \url{https://github.com/meaten/FlowChain-ICCV2023}
\end{abstract}

\section{Introduction}

Human trajectory prediction is a challenging problem because human movements are not deterministic,
unlike the billiard balls on a pool table.
Starting from deterministic approaches~\cite{helbing1995social, pellegrini2009you}, to tackle its indeterministic nature, many stochastic approaches are proposed, such as Gaussian distribution-based~\cite{alahi2016social, vemula2018social, mohamed2020social}, generative adversarial networks~(GANs)-based~\cite{gupta2018social, sadeghian2019sophie, dendorfer2020goal, dendorfer2021mg}, variational autoencoder~(VAE)-based~\cite{salzmann2020trajectron++, yuan2021agentformer, zhao2021you}, diffusion-based~\cite{gu2022stochastic}, and normalizing flow-based~\cite{scholler2021flomo} methods.
These methods can be applied to human-interactive  systems, including autonomous vehicles~\cite{deo2018convolutional, cui2019multimodal, li2019grip, wu2020motionnet, rudenko2020human} and social robots~\cite{liu2017human, butepage2018anticipating}. 

In these safety-critical applications, autonomous systems must infer the existence probability density distribution (``density'' for brevity) of other moving objects such as pedestrians around on all prediction steps for risk assessment (i.e., collision avoidance).
However, most of the indeterministic approaches mentioned above merely generate possible future trajectories and cannot estimate the density alone.
We need to further apply kernel density estimation (KDE) to these future trajectories for density estimation.

\begin{figure}
    \centering
    \includegraphics[width=\linewidth]{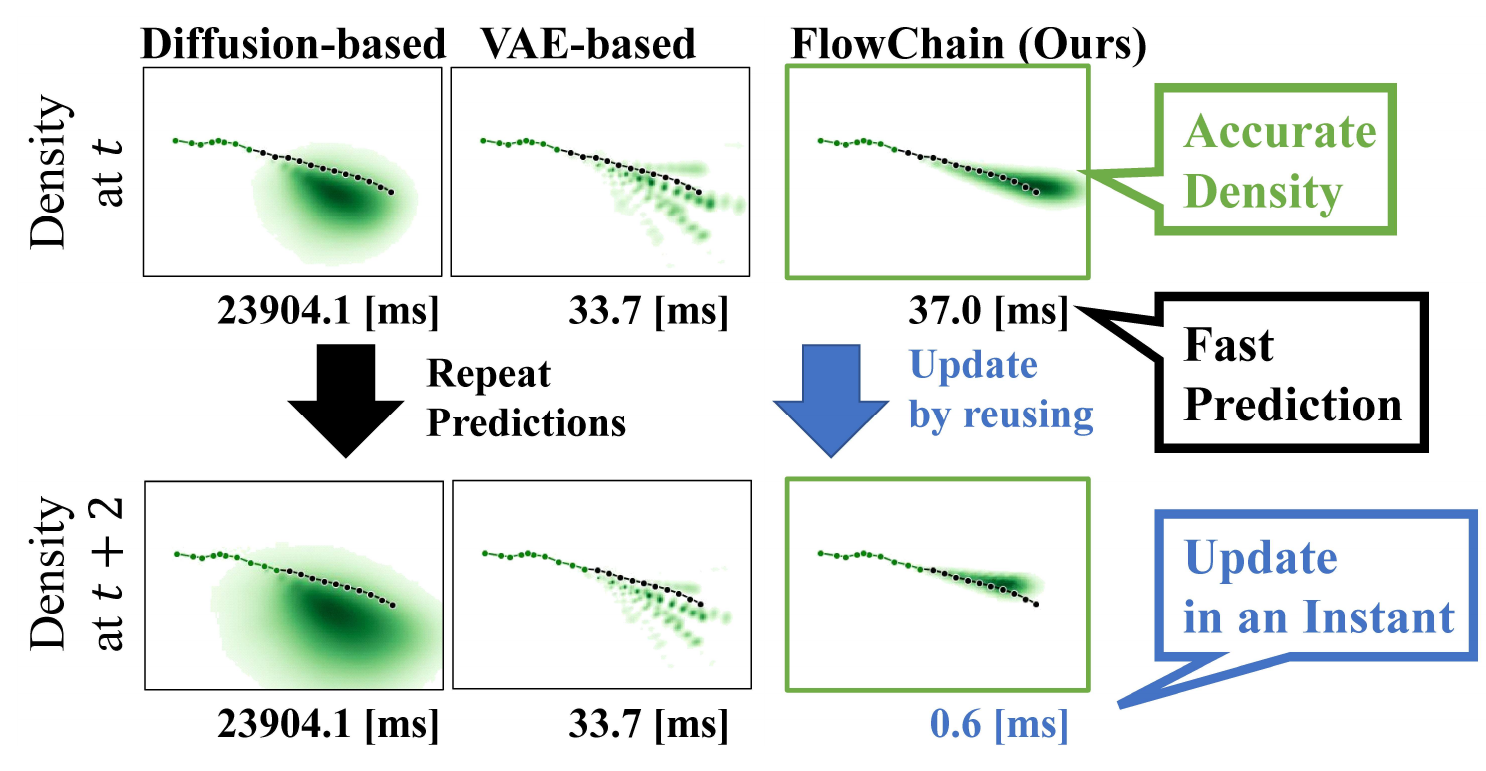}
    \caption{\textbf{Effects of our FlowChain.} Our FlowChain can rapidly estimate accurate density than VAE-based methods as shown in {\color[HTML]{00CC00} green}. Furthermore, we propose an update of estimated density that only requires an instant of time, as shown in \textcolor{blue}{blue}.}
    \label{fig:teaser}
\end{figure}

Thus, trajectory prediction with density estimation usually takes a larger computational cost
than only generating possible trajectories.
This computational cost accumulates upon every single prediction update along with highly-frequent observations.
This is unacceptable to autonomous systems that work in rapidly changing environments.

To address both computational cost and density estimation, we propose a new normalizing flow-based model named FlowChain (Fig.~\ref{fig:teaser}).
This speed and accuracy are because normalizing flows need no additional approximation such as Gaussian mixture modeling or KDE with limited samples.
FlowChain can estimate a density on each prediction time step thanks to our stacked normalizing flow architecture unlike  Flomo~\cite{scholler2021flomo}, which is closest to FlowChain. While Flomo uses one flow module that 
 only estimates the spatial density of a trajectory, FlowChain further predicts its spatio-temporal density so that the spatial density of an agent position at each time step is also predicted.
This stacking architecture also allows a rapid update of an estimated density in an instant of time \eg less than one millisecond.
This update is achieved by reusing the flow transformations and its log-det-jacobians that represent the \textit{motion trend} such as going straight or turning right. 
This update with the reuse significantly omits the computational cost while maintaining the accuracy of density estimation because the \textit{motion trend} doesn't change for several time steps.

Our contributions are three-fold as follows:
\begin{itemize}
    \item We propose a normalizing flow-based trajectory prediction network named FlowChain that rapidly estimates the accurate density of future positions on each prediction time step.
    \item Our update procedure for the density requires less than one millisecond by adopting new observations and reusing the \textit{motion trend} as the flow transformations.
    \item Our FlowChain achieved comparable prediction accuracy to other computationally-intense state-of-the-art methods on several benchmarks.
\end{itemize}

\section{Related Work}
\label{sec:related_work}
\noindent \textbf{Human Trajectory Prediction.}
The earliest works on human trajectory prediction~\cite{helbing1995social, pellegrini2009you} are deterministic regression models using attractive and repulsive forces between humans and goal positions.
However, human movements are indeterministic in nature.
Therefore, many approaches are proposed to model this uncertainty.
Social-LSTM~\cite{alahi2016social} and Social-STGCNN~\cite{mohamed2020social} regress the parameters of the bivariate Gaussian distributions to model the uncertainty of future trajectories.
These approaches~\cite{alahi2016social, mohamed2020social} are simple and require low computational cost but cannot catch the future multi-modality we often see at road intersections.

To address this multi-modality, many stochastic prediction models are proposed.
Trajectron++~\cite{salzmann2020trajectron++}, AgentFormer~\cite{yuan2021agentformer}, and ExpertTraj~\cite{zhao2021you} combined the aforementioned Gaussian regression with VAE~\cite{kingma2013auto}.
These models predict a bivariate Gaussian mixture based on stochastically sampled latent variables\cite{huang2019uncertainty, DBLP:conf/iccv/RhinehartMKL19}.
We can obtain the density by summing up each estimated Gaussian density map.
However, these Gaussian mixture methods have limited expressive power and thus predict inaccurate distributions.

On the other hand, GAN~\cite{goodfellow2020generative}-based approaches implicitly handle this uncertainty.
Social-GAN~\cite{gupta2018social}, SoPhie~\cite{sadeghian2019sophie}, Goal-GAN~\cite{dendorfer2020goal}, and MG-GAN~\cite{dendorfer2021mg} use a GAN architecture to predict diverse trajectories conditioned on random noises.
While they cannot estimate the density alone, we can estimate the density by applying KDE~\cite{rosenblatt1956remarks, parzen1962estimation} on sampled trajectories

The denoising diffusion probabilistic model~\cite{ho2020denoising} can also implicitly handle uncertainty by generating multiple trajectories.
Recently, denoising diffusion attracts much attention for its state-of-the-art performances in several generative tasks, including image synthesis~\cite{dhariwal2021diffusion} and audio generation~\cite{DBLP:conf/iclr/KongPHZC21}.
MID~\cite{gu2022stochastic} is a trajectory prediction model incorporating denoising diffusion, which achieves state-of-the-art performances.
We can also estimate the density of MID by applying KDE.
However, GAN-based and diffusion-based approaches cannot estimate accurate densities because KDE is performed on limited samples.

On the contrary, normalizing flow~\cite{tabak2010density, tabak2013family,rezende2015variational}-based trajectory prediction method named Flomo~\cite{scholler2021flomo} can rapidly generate future trajectories and their accurate probabilities thanks to the analytical computation of normalizing flows.
However, this approach only estimates the probability of each future trajectory and thus cannot estimate the density of positions on each prediction step alone.

\noindent \textbf{Normalizing Flows.}
Thanks to its bijective process, normalizing flow can compute the density analytically.
This analytical computation allows us fast and accurate density estimation because it does not need additional approximation such as KDE with limited samples.
For more expressive power, several bijective families were proposed such as affine coupling layers~\cite{dinh2015nice,dinh2017density}, autoregressive maps~\cite{germain2015made, papamakarios2017masked}, invertible linear transformations~\cite{kingma2018glow}, and invertible ResNet block~\cite{behrmann2019invertible, chen2019residual}.
All flows mentioned above have a topology constraint between the base and estimated densities.
This topology constraint limits the expressive power of normalizing flows.
To address this constraint, some approaches incorporated stochasticity into flows~\cite{wu2020stochastic}.
However, this stochasticity makes analytical density calculation impossible.
Therefore, further density approximation such as KDE is required for this stochastic normalizing flow. 
Other approaches~\cite{chen2020vflow, cornish2020relaxing, nielsen2020survae} use surjective layers, which relax the topology constraint but make analytical density computation with a little looseness.
In summary, the optimal normalizing flow model architecture for future trajectory density estimation is still unknown.

\noindent \textbf{Density Estimation.}
Kernel density estimation~\cite{rosenblatt1956remarks, parzen1962estimation} is the most representative non-parametric method that estimates a continuous density from a finite set of data.
Suppose that we have \(A\) input data and \(B\) points evaluated, the computational cost is \(\mathcal{O}(AB)\).
This is a huge cost for autonomous systems.
Therefore, several accelerations are proposed such as data binning~\cite{wand1994fast, gramacki2017fft, silverman1982algorithm, scott1985averaged}, fast sum updating~\cite{gasser1989discussion, seifert1994fast, fan1994fast, langrene2019fast}, fast Gauss transform~\cite{greengard1991fast, greengard1998new, lambert1999efficient}, and dual-tree method~\cite{gray2000n, gray2003nonparametric, lee2005dual}.
However, we confirmed that the KDE implementation utilizing GPU is faster than these accelerations including~\cite{bernacchia2011self, o2014reducing, o2016fast}.
We use the KDE utilizing GPU for comparison in Sec.~\ref{sec:exp}.

\begin{figure*}[t]
    \centering
    \includegraphics[width=\textwidth]{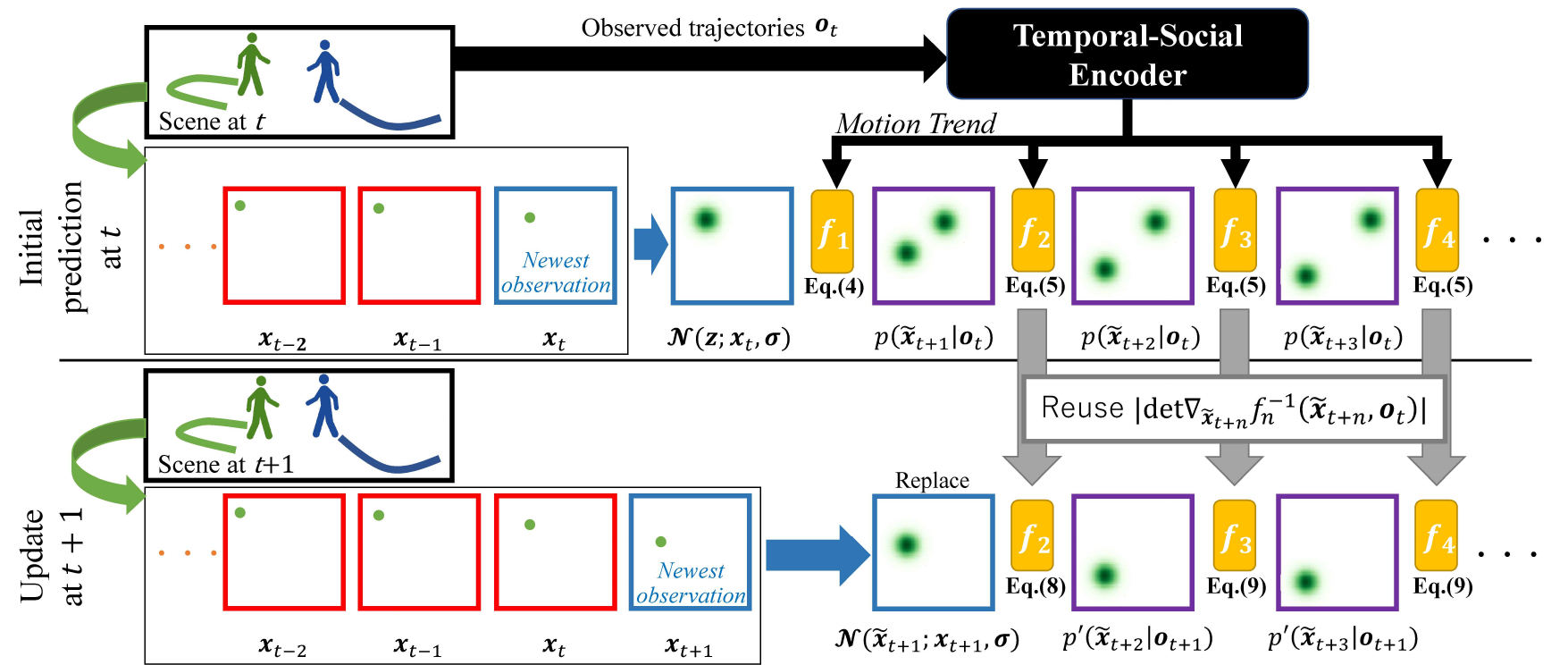}
    \caption{\textbf{Overview of our FlowChain model.}
    FlowChain consists of sequential conditional CIFs denoted by \(f_n\).
    \(f_{n}\) is conditioned on the feature vector denoted by \textit{Motion Trend} from the Temporal-Social Encoder, which encodes the observed trajectories \(\mathbf{o}_t\).
    Since \(\mathbf{o}_t\) includes the trajectories of other agents as well as an agent of interest (i.e, the target of prediction), the future trajectory is predicted while taking into account social interaction with the others.
    \(f_{n}\) transforms the density of the former time step \(p(\Tilde{\mathbf{x}}_{t+n-1}|\mathbf{o}_t)\) to the density of the next time step \(p(\Tilde{\mathbf{x}}_{t+n}|\mathbf{o}_t)\) as shown in the upper side.
    However, for the first step, the Gaussian distribution \(\mathcal{N}(\cdot; \mathbf{x}_t, \mathbf{\sigma})\), which is created from the \textit{newest observed position} \(\mathbf{x}_t\), is transformed.
    The boxes of past, current (i.e., newest), and future densities are colored by \textcolor{red}{red}, {\color[HTML]{87CEEB} blue}, and {\color[HTML]{800080}purple}, respectively.
    In the update procedure shown on the bottom side, the Gaussian distribution \(\mathcal{N}(\cdot; \mathbf{x}_{n+1}, \mathbf{\sigma})\), created from \textit{newest observed position} \(\mathbf{x}_{t+1}\) at \(t + 1\), replace the estimated density from the former time step.
    The other calculations are reused as shown by \textcolor{gray}{gray arrows}.
    We can properly update the density by using the \textit{newest observed position} at $t+1$, \(\mathbf{x}_{t+1}\), instead of \(p(\Tilde{\mathbf{x}}_{t+1}|\mathbf{o}_t)\).
    In a toy example shown in this figure, another incorrect mode observed in the predicted densities, \(p(\Tilde{\mathbf{x}}_{t+2}|\mathbf{o}_t)\) and \(p(\Tilde{\mathbf{x}}_{t+3}|\mathbf{o}_t)\), is successfully suppressed in the updated predictions \(p'(\Tilde{\mathbf{x}}_{t+2}|\mathbf{o}_{t+1})\) and \(p'(\Tilde{\mathbf{x}}_{t+3}|\mathbf{o}_{t+1})\).
    As shown in Sec.~\ref{sec:exp}, our experiments validate this advantage of our proposed fast update procedure is greater than its disadvantage (i.e., a decrease in the prediction accuracy due to the reuse of the motion trend) so that the updated predictions are better than the initial predictions (e.g., \(p'(\Tilde{\mathbf{x}}_{t+3}|\mathbf{o}_{t+1})\) is better than \(p(\Tilde{\mathbf{x}}_{t+3}|\mathbf{o}_t)\)in the figure).
    }
    \label{fig:overview}
\end{figure*}

\section{Proposed Approach}
In this section, we present our normalizing flow-based trajectory prediction named FlowChain and its update of estimated density.
First, we formulate the trajectory prediction task and conditional CIFs in Sec.~\ref{subsec:problem_formulation}. and Sec.~\ref{subsec:normalizing_flow} respectively.
Then, we present our FlowChain model architecture in Sec.~\ref{subsec:flowchain}.
After that, we describe its fast update procedure of estimated density in Sec.\ref{subsec:update}.
Finally, we explain how to obtain the density map in Sec.~\ref{subsection:density}.

\subsection{Problem Formulation}
\label{subsec:problem_formulation}
We aim to estimate agents' plausible future trajectory densities from their past trajectories.
The trajectory of each agent is defined as a sequence \((\mathbf{x}_0, \mathbf{x}_1, \dots, \mathbf{x}_T)\) of positions \(\mathbf{x}_t = (x_t, y_t)\) over discrete timesteps \(t \in \{0, \dots, T\}\).
Given the current time step $t$, we predict future \(T_{\rm f}\)-step trajectory densities \(\{p(\Tilde{\mathbf{x}}^i_{t+1}|\mathbf{o}_t), \dots, p(\Tilde{\mathbf{x}}^i_{t+T_{\rm f}}|\mathbf{o}_t)\}\) given \(T_{\rm o}\)-step observed trajectory \(\mathbf{o}_t = \{(\mathbf{x}^i_{t-T_{\rm o}-1}, \dots, \mathbf{x}^i_t)| \quad i \in \{1, 2, \dots, N_{a}\}\}\), where $N_{a}$ and \(i\) denote the number of agents and the index of each agent, respectively.
This index will be omitted for brevity in later use.
We denote \(\mathbf{x}\) as an observed position, \(\Tilde{\mathbf{x}}\) as a predicted future position, and \(\hat{\mathbf{x}}\) as a ground truth future position.

\subsection{Conditional Continuously-indexed Flow}
\label{subsec:normalizing_flow}

Conditional normalizing flow is a density estimation method that uses a parameterized bijective transformation \(f\) conditioned on additional information \(\mathbf{c}\) such as an encoded vector from observed trajectories, i.e., \(\mathbf{o}_{t}\), for trajectory prediction.
The bijection \(f\) transforms a base density \(p(\mathbf{z})\) to an expressive density \(p(\mathbf{y})\) by the analytical computation of the change-of-variables formula as follows:
\begin{align}
    \mathbf{y} &= f(\mathbf{z}) \\
    p(\mathbf{y}) &= p(\mathbf{z})|{\rm det}\nabla_\mathbf{y}f^{-1}(\mathbf{y}, \mathbf{c})|
\end{align}
The parameters of bijective transformation \(f\) can be learned by maximizing the likelihood of samples \(\hat{\mathbf{y}}\) from datasets, or minimizing the negative log-likelihood as follows:
\begin{align}
    NLL = - \log p(\hat{\mathbf{y}}|\mathbf{c})
    \label{eq:nll}
\end{align}

A normalizing flow with only bijective transformations has a constraint that the topology of a base density is preserved.
For example, if the base density is a simple unimodal Gaussian distribution, the normalizing flow fails to fit a distinct multi-modal density.
To avoid this problem of bijective transformation, we use continuously-indexed flows (CIFs)~\cite{cornish2020relaxing}, akin to an infinite mixture of normalizing flows.
While CIFs potentially avoid the topology constraint, CIFs still enable analytical density computation with a little looseness \(\mathcal{E}\) because CIFs incorporate inference and generative surjections according to the survey~\cite{nielsen2020survae}.
We investigated the tradeoff between the expressive power and exactness of estimated density on CIFs.
We concluded this looseness of estimated density can be ignored in trajectory prediction, as we show in the supplemental materials.
Based on the above discussion, We use conditional CIFs as the base normalizing flow model of our FlowChain.

\subsection{FlowChain model}
\label{subsec:flowchain}

The overview of the FlowChain model is shown in Fig.~\ref{fig:overview}.
Given the current time step \(t\), we estimate future trajectory densities \(\{p(\Tilde{\mathbf{x}}_{t+1}|\mathbf{o}_t), \dots, p(\Tilde{\mathbf{x}}_{t+T_{\rm f}}|\mathbf{o}_t)\}\) by chaining the conditional CIFs, each of which transforms the density of the former time step \(p(\Tilde{\mathbf{x}}_{t+n-1}|\mathbf{o}_t)\) to the density of the next time step \(p(\Tilde{\mathbf{x}}_{t+n}|\mathbf{o}_t)\), shown in the upper side of Fig.~\ref{fig:overview}.
Our formulation consists of two equations: the initial condition for the first step prediction (Eq.(\ref{eq:first_step})) and the recurrence relation between the latter steps (Eq.(\ref{eq:recurrence_relation})) as follows.
\begin{align}
    p(\Tilde{\mathbf{x}}_{t+1}|\mathbf{o}_t) &= \mathcal{N}(\mathbf{z}; \mathbf{x}_t, \mathbf{\sigma})|{\rm det}\nabla_{\Tilde{\mathbf{x}}_{t+1}} f_1^{-1}(\Tilde{\mathbf{x}}_{t+1}, \mathbf{o}_t)| \label{eq:first_step}\\
    p(\Tilde{\mathbf{x}}_{t+n}|\mathbf{o}_t) &= p(\Tilde{\mathbf{x}}_{t+n-1}|\mathbf{o}_t)|{\rm det}\nabla_{\Tilde{\mathbf{x}}_{t+n}}f_n^{-1}(\Tilde{\mathbf{x}}_{t+n}, \mathbf{o}_t)| \label{eq:recurrence_relation}\\
    &= \exp \bigg( \log \mathcal{N}(\mathbf{z}; \mathbf{x}_t, \mathbf{\sigma}) \nonumber\\
    & \quad \quad + \sum_{n'=1}^{n}\log|{\rm det}\nabla_{\Tilde{\mathbf{x}}_{t+n'}}f_{n'}^{-1}(\Tilde{\mathbf{x}}_{t+n'}, \mathbf{o}_t)|\bigg) \label{eq:density}
\end{align}
where \(n \in \{2, 3, \dots, T_{\rm f}\}\). 
\(f_n\) is a conditional CIF that transforms the sample \(\Tilde{\mathbf{x}}_{t+n-1}\) from a former density \(p(\Tilde{\mathbf{x}}_{t+n-1}|\mathbf{o}_t)\) to the sample \(\Tilde{\mathbf{x}}_{t+n}\) from a next density \(p(\Tilde{\mathbf{x}}_{t+n}|\mathbf{o}_t)\) as \(\Tilde{\mathbf{x}}_{t+n} = f_n(\Tilde{\mathbf{x}}_{t+n-1}, \mathbf{o}_t)\).
\(f_1\) transforms the sample \(\mathbf{z}\) from a base Gaussian density \(\mathcal{N}(\cdot; \mathbf{x}_t, \mathbf{\sigma})\).
\(\mathcal{N}(\cdot; \mathbf{x}_t, \mathbf{\sigma})\) is created from the \textit{newest observed position} \(\mathbf{x}_t\) as a mean position.
The standard deviation \(\sigma\) is a trainable parameter of FlowChain model.
All of $f_{1}$ and $f_{n}$ are independently trained with the loss function (Eq.(\ref{eq:nll})) in our method.

From \(f_n\), we can sample multiple trajectories for best-of-N metrics~\cite{gupta2018social} by sampling multiple \(\mathbf{z}\) as follows:
\begin{align}
    \{\Tilde{\mathbf{x}}_{t+1}, \dots, \Tilde{\mathbf{x}}_{t+n}\} = \{f_1(\mathbf{z},\mathbf{o}_t), \dots, f_n \circ \cdots \circ f_2 \circ f_1(\mathbf{z},\mathbf{o}_t)\} \label{eq:generated_trajectories}
\end{align}



%

\subsection{Update of estimated density}
\label{subsec:update}

For accurate prediction by reusing the previous predictions, we focused on two features of the observed trajectory: the \textit{newest observed position} and \textit{\textit{motion trend}}, shown in Fig.~\ref{fig:overview}.
Our FlowChain generates future trajectories by transforming the \textit{newest observed position} \(\mathbf{x}_t\) based on the \textit{motion trend} \eg go straight or turning left/right, modeled by each flow \(f_n(\mathbf{x}_{t+n}, \mathbf{o}_t)\).
We cannot reuse the \textit{newest observed position} \(\mathbf{x}_t\) at \(t\) for the next time step because this position is a starting point of the future trajectories and thus greatly affect the accuracy of probability density estimation.
Therefore, FlowChain adopt the \textit{newest observed position} \(\mathbf{x}_{t+n}\) at \(t+n\) by replacing the estimated density \(p(\Tilde{\mathbf{x}}_{t+n}|\mathbf{o}_t)\) with the Gaussian density \(\mathcal{N}(\cdot; \mathbf{x}_{t+n}, \mathbf{\sigma})\), as shown in the bottom of Fig.~\ref{fig:overview}. This replacement propagates the information of the \textit{newest observed position} properly to the latter time steps.

\begin{figure*}[t]
    \centering
    \includegraphics[width=0.95\textwidth]{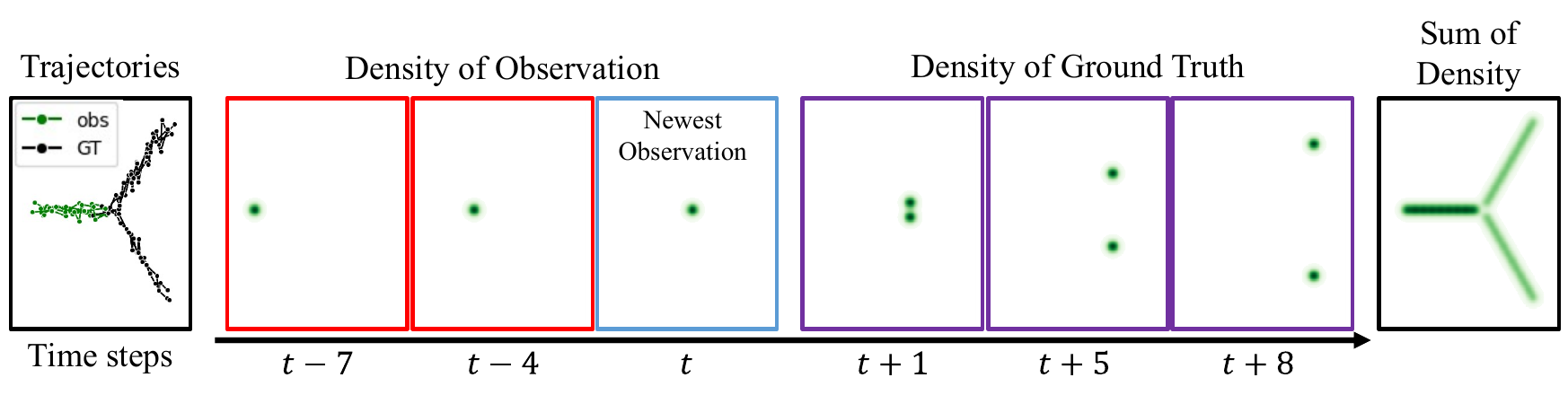}
    \caption{\textbf{Overview of \textit{Simfork} dataset.} The trajectories shown in the left figure are examples of the observed and ground truth trajectories for training. The position of each step is sampled from the distributions of the middle figures. The sum of each distribution is shown in the right figure. The boxes of past, current (i.e., newest), and future densities are colored by \textcolor{red}{red}, {\color[HTML]{87CEEB} blue}, and {\color[HTML]{800080}purple}, respectively.}
    \label{fig:simfork}
\end{figure*}

On the other hand, the \textit{motion trend} can be reused because it doesn't change for several time steps unless a sudden strong external force is applied to a prediction target.
Therefore, our FlowChain model reuses its trend in each flow transformation  \(f_n(\Tilde{\mathbf{x}}_{t+n}, \mathbf{o}_t)\).
Specifically, we use the previously generated trajectories \(\{\Tilde{\mathbf{x}}_{t+1}, \dots, \Tilde{\mathbf{x}}_{t+n}\}\).
Thus, we can reuse the corresponding log-det-jacobians \(\log|{\rm det}\nabla_{\Tilde{\mathbf{x}}_{t+n}}f_{n}^{-1}(\Tilde{\mathbf{x}}_{t+n+1}, \mathbf{o}_t)|\) which is the variation of the estimated densities, as shown by the \textcolor{gray}{gray arrows} in Fig.~\ref{fig:overview}.
The equation of FlowChain's update is expressed similar to Eq.(\ref{eq:first_step}) and Eq.(\ref{eq:recurrence_relation}) as follows:
\begin{align}
    p'(\Tilde{\mathbf{x}}_{t+2}|\mathbf{o}_{t+1}) &= \mathcal{N}(\Tilde{\mathbf{x}}_{t+1}; \mathbf{x}_{t+1}, \mathbf{\sigma})|{\rm det}\nabla_{\Tilde{\mathbf{x}}_{t+2}}f_2^{-1}(\Tilde{\mathbf{x}}_{t+2}, \mathbf{o}_t)| \\
    p'(\Tilde{\mathbf{x}}_{t+n}|\mathbf{o}_{t+1}) &= p'(\Tilde{\mathbf{x}}_{t+n-1}|\mathbf{o}_{t+1})|{\rm det}\nabla_{\Tilde{\mathbf{x}}_{t+n}}f_n^{-1}(\Tilde{\mathbf{x}}_{t+n})| \\
    &= \exp \bigg( \log \mathcal{N}(\Tilde{\mathbf{x}}_{t+1}; \mathbf{x}_{t+1}, \mathbf{\sigma}) \nonumber\\
    & \quad \quad + \sum_{n'=2}^{n}\log|{\rm det}\nabla_{\Tilde{\mathbf{x}}_{t+n'}}f_{n'}^{-1}(\Tilde{\mathbf{x}}_{t+n'}, \mathbf{o}_t)|\bigg) 
\end{align}
where, \(n \in \{3, 4, \dots, T_{\rm f}\} \).
Thanks to reusing each flow transformation \(f_n(\mathbf{x}_{t+n}, \mathbf{o}_t)\), we don't need to evaluate log-det-jacobians \(\log|{\rm det}\nabla_{\Tilde{\mathbf{x}}_{t+n}}f_n^{-1}(\Tilde{\mathbf{x}}_{t+n})|\) again.
This will greatly omit the computational cost.
As a result, we can update the probability density by only calculating the Gaussian base distribution, summing up the log-det-jacobians, and taking exponentials, which take less than 1\textit{ms} in total.
This rapid update is beneficial, especially for autonomous systems where computational resources are limited and often become busy by executing other computationally-intense tasks such as object detection, mapping, and path planning.

\begin{table*}[t]
\centering
\caption{\textbf{Quantitative results of trajectory prediction on \textit{ETH/UCY} dataset with Best-of-20 metrics.} T + I denotes its method takes a trajectory and an image as inputs while T denotes its method takes only a trajectory. Scores are in meters. {\color[HTML]{fe0000} \textbf{Red}} and {\color[HTML]{0000ee} \textbf{blue}} scores denote the best and the second-best in methods with T.  Lower is better. The results of Trajectron++~\cite{salzmann2020trajectron++} and MID~\cite{gu2022stochastic} are updated according to an implementation issue \#53 on the Trajectron++ GitHub page.}

\begin{tabular}{ll|rrrrrrrrrr|rr}
\hline
 & \multicolumn{1}{c|}{} & \multicolumn{2}{c}{\textbf{ETH}} & \multicolumn{2}{c}{\textbf{HOTEL}} & \multicolumn{2}{c}{\textbf{UNIV}} & \multicolumn{2}{c}{\textbf{ZARA1}} & \multicolumn{2}{c|}{\textbf{ZARA2}} & \multicolumn{2}{c}{\textbf{Mean}} \\ \cline{3-14} 
\multirow{-2}{*}{\textbf{Method}} & \multicolumn{1}{c|}{\multirow{-2}{*}{\textbf{Input}}} & \multicolumn{1}{c}{ADE} & \multicolumn{1}{c}{\cellcolor[HTML]{EFEFEF}FDE} & \multicolumn{1}{c}{ADE} & \multicolumn{1}{c}{\cellcolor[HTML]{EFEFEF}FDE} & \multicolumn{1}{c}{ADE} & \multicolumn{1}{c}{\cellcolor[HTML]{EFEFEF}FDE} & \multicolumn{1}{c}{ADE} & \multicolumn{1}{c}{\cellcolor[HTML]{EFEFEF}FDE} & \multicolumn{1}{c}{ADE} & \multicolumn{1}{c|}{\cellcolor[HTML]{EFEFEF}FDE} & \multicolumn{1}{c}{ADE} & \multicolumn{1}{c}{\cellcolor[HTML]{EFEFEF}FDE} \\ \hline
SoPhie\cite{sadeghian2019sophie} & T + I & 0.70 & \cellcolor[HTML]{EFEFEF}1.43 & 0.76 & \cellcolor[HTML]{EFEFEF}1.67 & 0.54 & \cellcolor[HTML]{EFEFEF}1.24 & 0.30 & \cellcolor[HTML]{EFEFEF}0.63 & 0.38 & \cellcolor[HTML]{EFEFEF}0.78 & 0.54 & \cellcolor[HTML]{EFEFEF}1.15 \\
Goal-GAN\cite{dendorfer2020goal} & T + I & 0.59 & \cellcolor[HTML]{EFEFEF}1.18 & 0.19 & \cellcolor[HTML]{EFEFEF}0.35 & 0.60 & \cellcolor[HTML]{EFEFEF}1.19 & 0.43 & \cellcolor[HTML]{EFEFEF}0.87 & 0.32 & \cellcolor[HTML]{EFEFEF}0.65 & 0.43 & \cellcolor[HTML]{EFEFEF}0.85 \\
MG-GAN\cite{dendorfer2021mg} & T + I & 0.47 & \cellcolor[HTML]{EFEFEF}0.91 & 0.14 & \cellcolor[HTML]{EFEFEF}0.24 & 0.54 & \cellcolor[HTML]{EFEFEF}1.07 & 0.36 & \cellcolor[HTML]{EFEFEF}0.73 & 0.29 & \cellcolor[HTML]{EFEFEF}0.60 & 0.36 & \cellcolor[HTML]{EFEFEF}0.71 \\ \hline
Social-LSTM\cite{alahi2016social} & T & 1.09 & \cellcolor[HTML]{EFEFEF}2.35 & 0.79 & \cellcolor[HTML]{EFEFEF}1.76 & 0.67 & \cellcolor[HTML]{EFEFEF}1.40 & 0.47 & \cellcolor[HTML]{EFEFEF}1.00 & 0.56 & \cellcolor[HTML]{EFEFEF}1.17 & 0.72 & \cellcolor[HTML]{EFEFEF}1.54 \\
Social-GAN\cite{gupta2018social} & T & 0.87 & \cellcolor[HTML]{EFEFEF}1.62 & 0.67 & \cellcolor[HTML]{EFEFEF}1.37 & 0.76 & \cellcolor[HTML]{EFEFEF}1.52 & 0.35 & \cellcolor[HTML]{EFEFEF}0.68 & 0.42 & \cellcolor[HTML]{EFEFEF}0.84 & 0.61 & \cellcolor[HTML]{EFEFEF}1.21 \\
STGAT\cite{huang2019stgat} & T & 0.65 & \cellcolor[HTML]{EFEFEF}1.12 & 0.35 & \cellcolor[HTML]{EFEFEF}0.66 & 0.52 & \cellcolor[HTML]{EFEFEF}1.10 & 0.34 & \cellcolor[HTML]{EFEFEF}0.69 & 0.29 & \cellcolor[HTML]{EFEFEF}0.60 & 0.43 & \cellcolor[HTML]{EFEFEF}0.83 \\
Social-STGCNN\cite{mohamed2020social} & T & 0.64 & \cellcolor[HTML]{EFEFEF}1.11 & 0.49 & \cellcolor[HTML]{EFEFEF}0.85 & 0.44 & \cellcolor[HTML]{EFEFEF}0.79 & 0.34 & \cellcolor[HTML]{EFEFEF}0.53 & 0.30 & \cellcolor[HTML]{EFEFEF}0.48 & 0.44 & \cellcolor[HTML]{EFEFEF}0.75 \\
PECNet\cite{mangalam2020not} & T & 0.61 & \cellcolor[HTML]{EFEFEF}1.07 & {\color[HTML]{0000EE} \textbf{0.22}} & \cellcolor[HTML]{EFEFEF}0.39 & 0.34 & \cellcolor[HTML]{EFEFEF}0.56 & 0.25 & \cellcolor[HTML]{EFEFEF}0.45 & {\color[HTML]{0000EE} \textbf{0.19}} & \cellcolor[HTML]{EFEFEF}{\color[HTML]{0000EE} \textbf{0.33}} & 0.32 & \cellcolor[HTML]{EFEFEF}0.56 \\
Trajectron++\cite{salzmann2020trajectron++} & T & {\color[HTML]{0000EE} \textbf{0.61}} & \cellcolor[HTML]{EFEFEF}1.03 & {\color[HTML]{FE0000} \textbf{0.20}} & \cellcolor[HTML]{EFEFEF}{\color[HTML]{FE0000} \textbf{0.28}} & {\color[HTML]{0000EE} \textbf{0.30}} & \cellcolor[HTML]{EFEFEF}{\color[HTML]{0000EE} \textbf{0.55}} & {\color[HTML]{0000EE} \textbf{0.24}} & \cellcolor[HTML]{EFEFEF}{\color[HTML]{0000EE} \textbf{0.41}} & {\color[HTML]{FE0000} \textbf{0.18}} & \cellcolor[HTML]{EFEFEF}{\color[HTML]{FE0000} \textbf{0.32}} & {\color[HTML]{0000EE} \textbf{0.31}} & \cellcolor[HTML]{EFEFEF}{\color[HTML]{FE0000} \textbf{0.52}} \\
MID\cite{gu2022stochastic} & T & {\color[HTML]{FE0000} \textbf{0.55}} & \cellcolor[HTML]{EFEFEF}{\color[HTML]{FE0000} \textbf{0.88}} & {\color[HTML]{FE0000} \textbf{0.20}} & \cellcolor[HTML]{EFEFEF}{\color[HTML]{0000EE} \textbf{0.35}} & {\color[HTML]{0000EE} \textbf{0.30}} & \cellcolor[HTML]{EFEFEF}{\color[HTML]{0000EE} \textbf{0.55}} & 0.29 & \cellcolor[HTML]{EFEFEF}0.51 & 0.20 & \cellcolor[HTML]{EFEFEF}0.38 & {\color[HTML]{0000EE} \textbf{0.31}} & \cellcolor[HTML]{EFEFEF}{\color[HTML]{0000EE} \textbf{0.53}} \\
Social Implicit\cite{mohamed2022social} & T & 0.66 & \cellcolor[HTML]{EFEFEF}1.44 & {\color[HTML]{FE0000} \textbf{0.20}} & \cellcolor[HTML]{EFEFEF}0.36 & 0.31 & \cellcolor[HTML]{EFEFEF}0.60 & 0.25 & \cellcolor[HTML]{EFEFEF}0.50 & 0.22 & \cellcolor[HTML]{EFEFEF}0.43 & 0.33 & \cellcolor[HTML]{EFEFEF}0.67 \\
FlowChain(Ours) & T & {\color[HTML]{FE0000} \textbf{0.55}} & \cellcolor[HTML]{EFEFEF}{\color[HTML]{0000EE} \textbf{0.99}} & {\color[HTML]{FE0000} \textbf{0.20}} & \cellcolor[HTML]{EFEFEF}{\color[HTML]{0000EE} \textbf{0.35}} & {\color[HTML]{FE0000} \textbf{0.29}} & \cellcolor[HTML]{EFEFEF}{\color[HTML]{FE0000} \textbf{0.54}} & {\color[HTML]{FE0000} \textbf{0.22}} & \cellcolor[HTML]{EFEFEF}{\color[HTML]{FE0000} \textbf{0.40}} & 0.20 & \cellcolor[HTML]{EFEFEF}0.34 & {\color[HTML]{FE0000} \textbf{0.29}} & \cellcolor[HTML]{EFEFEF}{\color[HTML]{FE0000} \textbf{0.52}} \\ \hline
\end{tabular}
\label{tab:ethucy}
\end{table*}

\begin{table}[t]
\centering
\caption{\textbf{Quantitative results of trajectory prediction on \textit{SDD} dataset with Best-of-20 metrics.} 
Scores are in pixels.
See the caption of Tab.~\ref{tab:ethucy} also for details.
}
\begin{tabular}{ll|r
>{\columncolor[HTML]{EFEFEF}}r }
\hline
\textbf{Method} & \multicolumn{1}{c|}{\textbf{Input}} & \multicolumn{1}{c}{\textbf{ADE}} & \multicolumn{1}{c}{\cellcolor[HTML]{EFEFEF}\textbf{FDE}} \\ \hline
SoPhie\cite{sadeghian2019sophie} & T + I & 16.27 & 29.38 \\
Goal-GAN\cite{dendorfer2020goal} & T + I & 12.20 & 22.10 \\
MG-GAN\cite{dendorfer2021mg} & T + I & 13.60 & 25.80 \\ \hline
Social-LSTM\cite{alahi2016social} & T & 31.19 & 56.97 \\
Social-GAN\cite{gupta2018social} & T & 27.25 & 41.44 \\
STGAT\cite{huang2019stgat} & T & \multicolumn{1}{l}{14.85} & \multicolumn{1}{l}{\cellcolor[HTML]{EFEFEF}28.17} \\
Social-STGCNN\cite{mohamed2020social} & T & \multicolumn{1}{l}{20.76} & \multicolumn{1}{l}{\cellcolor[HTML]{EFEFEF}33.18} \\
PECNet\cite{mangalam2020not} & T & {\color[HTML]{0000EE} \textbf{9.97}} & {\color[HTML]{FE0000} \textbf{15.89}} \\
Trajectron++\cite{salzmann2020trajectron++} & T & 11.40 & 20.12 \\
MID\cite{gu2022stochastic} & T & 10.31 & 17.37 \\
FlowChain (Ours) & T & {\color[HTML]{FE0000} \textbf{9.93}} & \cellcolor[HTML]{EFEFEF}{\color[HTML]{0000EE} \textbf{17.17}} \\ \hline
\end{tabular}
\label{tab:sdd}
\end{table}

\subsection{How to obtain the density map}
\label{subsection:density}

The density map, not like the mere one-point estimate, is needed when autonomous systems perform visualization of density or successive tasks such as navigation.
We can correct the possible future points with estimated density by sampling multiple trajectories from Eq.(\ref{eq:generated_trajectories}) and its density from Eq.(\ref{eq:density}).
The density map is obtained by collecting these points with density at the same time step.

\section{Experiments and Results Analysis}
\label{sec:exp}
\subsection{Experimental Setup}
\noindent \textbf{Datasets}.
Experiments were conducted on various datasets including our developed dataset with synthesized data \textit{Simfork}, and real-image benchmarks including \textit{ETH/UCY}~\cite{pellegrini2010improving, lerner2007crowds} and \textit{SDD}~\cite{robicquet2016learning}.
On all the datasets, the numbers of past observed and future predicted steps are \(T_o = 8\) and \(T_f = 12\), respectively.
All steps have an interval of 0.4 seconds in accordance with the literature such as~\cite{gupta2018social,salzmann2020trajectron++}.

\textit{Simfork} is a synthesized dataset for comparison on the accuracy of predicted densities, which are produced as explained in Sec.~\ref{subsection:density}.
In all trajectories in this dataset, either of the two types of traveling directions (i.e., turning left or right) is observed after a straight trajectory, as shown in Fig.~\ref{fig:simfork}.
Positions before and after the fork (colored by green and black, respectively, in the leftmost figure of Fig.~\ref{fig:simfork}) are used as past observed and future ground-truth positions, respectively.
Each position of a trajectory is independently perturbed by the Gaussian noise, which results in the density at each time step.

Two pedestrian trajectory datasets, \textit{ETH}~\cite{pellegrini2010improving} and \textit{UCY}~\cite{lerner2007crowds},  are merged to \textit{ETH/UCY}.
This dataset has five scenes: ETH, HOTEL, UNIV, ZARA1, and ZARA2.
For training and testing, we follow the standard leave-one-out approach, where we train on 4 scenes and test on the remaining one.
We follow the train-validation-test split of Social-GAN~\cite{gupta2018social}.

\textit{SDD} consists of 20 scenes captured in bird's eye view using aerial drones.
This dataset contains various moving agents such as pedestrians, bicycles, and cars, but we focus on pedestrians for comparison, as with Trajnet~\cite{sadeghian2018trajnet}.
We also follow the train-test split of Trajnet~\cite{sadeghian2018trajnet}.

\noindent\textbf{Metrics}
Evaluation of accuracy on trajectory prediction is done with the widely-used metrics, Average Displacement Error (ADE) and Final Displacement Error (FDE).
ADE is the average L2 distance between the predicted and ground truth trajectories over time.
FDE is the L2 distance between the endpoints of the predicted and ground truth trajectories.
We follow the Best-of-N procedure~\cite{gupta2018social} with \(N=20\).

For evaluation of the accuracy of density estimation, we use earth mover's distance (EMD) and log-probability.
EMD is a measure of the distance between two densities by the minimum cost to move one density pile to the other.
EMD is calculated between the estimated density and ground truth density on each prediction step.
Thus, we use EMD on \textit{Simfork} dataset, where the ground truth densities are available.
The advantage of EMD over the widely-used KL divergence is that EMD works properly even if the two density is disjointed while KL divergence returns a constant value.
We use log-probability on the pedestrian trajectory prediction datasets, where the ground truth density is unavailable.
The log-probability is calculated on the ground truth trajectory instead of density.

\noindent\textbf{Implementation Details}.
Each conditional CIFs in our FlowChain used three-layer RealNVP~\cite{dinh2017density} with multilayer perceptions (MLPs) of three hidden layers and 128 hidden units.
All the conditional CIFs are with independent weights conditioned on the feature vectors from Trajectron++\cite{salzmann2020trajectron++} encoder as the temporal-social encoder.
We used a batch size of 128 and ADAM optimizer~\cite{DBLP:journals/corr/KingmaB14} with a learning rate of \(10^{-4}\).
All experiments were conducted on a single NVIDIA A100 GPU.
We use a KDE implementation utilizing GPU as mentioned in Sec.~\ref{sec:related_work}.


\subsection{Prediction Accuracy}

A wide range of previous trajectory prediction methods is quantitatively compared with our FlowChain on \textit{ETH/UCY} and \textit{SDD} datasets, shown in Tab.~\ref{tab:ethucy} and Tab.~\ref{tab:sdd}.
We didn't include some methods~\cite{yu2020spatio, yuan2021agentformer, mangalam2021goals, sun2021three, yue2022human} due to a different train-validation-test split.

For \textit{ETH/UCY} dataset, FlowChain achieved mean ADE/FDE of \(0.29/0.52\) in meters, which are the best among all trajectory prediction methods.
%
On \textit{SDD} dataset, FlowChain achieved average ADE/FDE of \(9.93/17.17\) in pixels, which are the best and second-best among all methods taking only trajectories as inputs~\cite{alahi2016social,gupta2018social,salzmann2020trajectron++, mangalam2020not,sun2021three, gu2022stochastic}.

Although our focus is on density estimation with a minimal computational cost, FlowChain is comparable with the state-of-the-art methods also in terms of errors.
We didn't evaluate the accuracy of future trajectories from FlowChain's update because the trajectories do not change in the update as mentioned in Sec.~\ref{subsec:update}.

\subsection{Analysis on Density Estimation}

\begin{figure}[t]
    \centering
    \includegraphics[width=\linewidth]{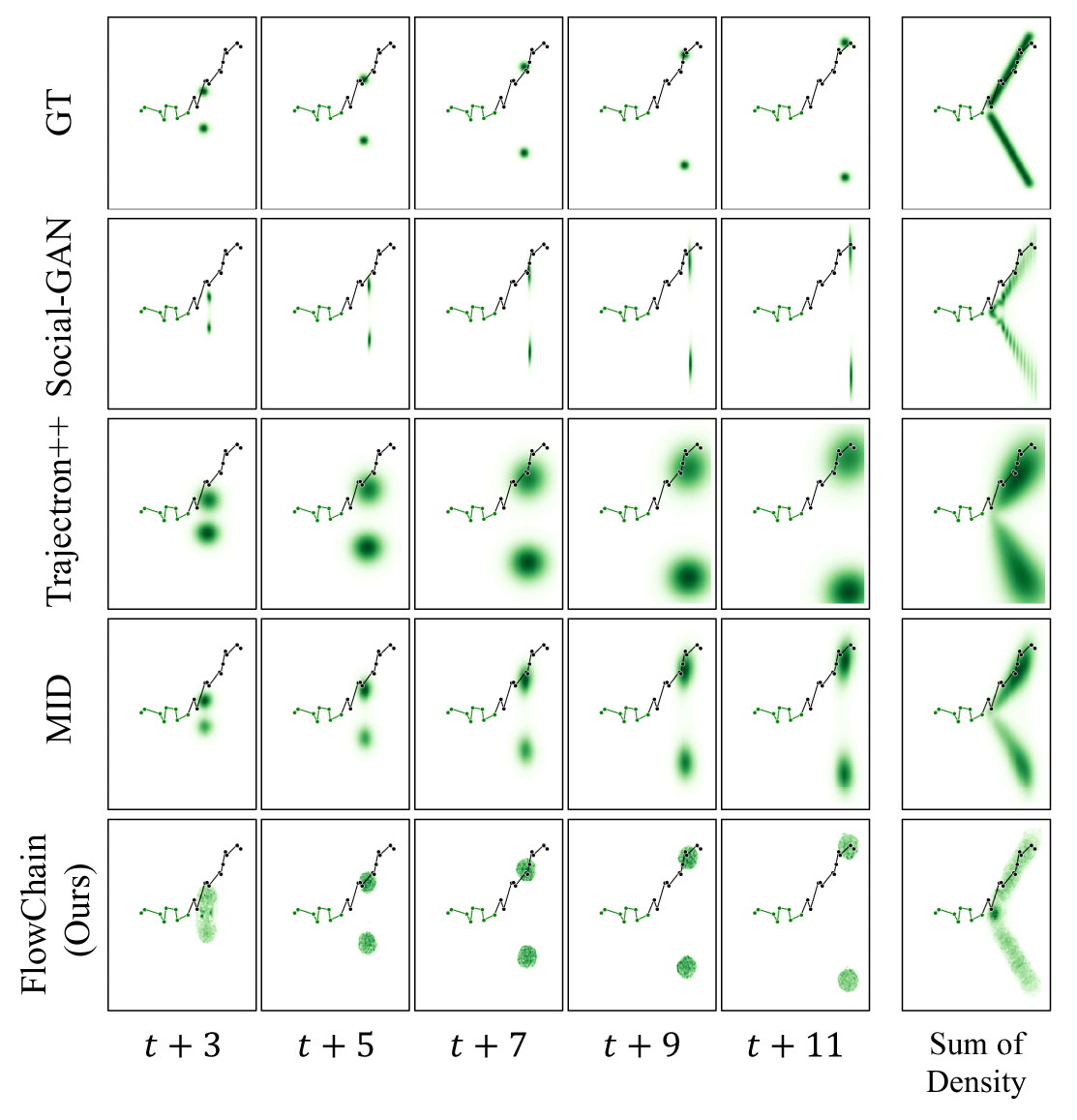}\\
    \vspace*{-2mm}
    \caption{
    \textbf{Temporal estimated densities on \textit{Simfork} dataset.}
    }
    \label{fig:exp_simfork}
\end{figure}

\begin{figure}[t]
    \centering
    \includegraphics[width=\linewidth]{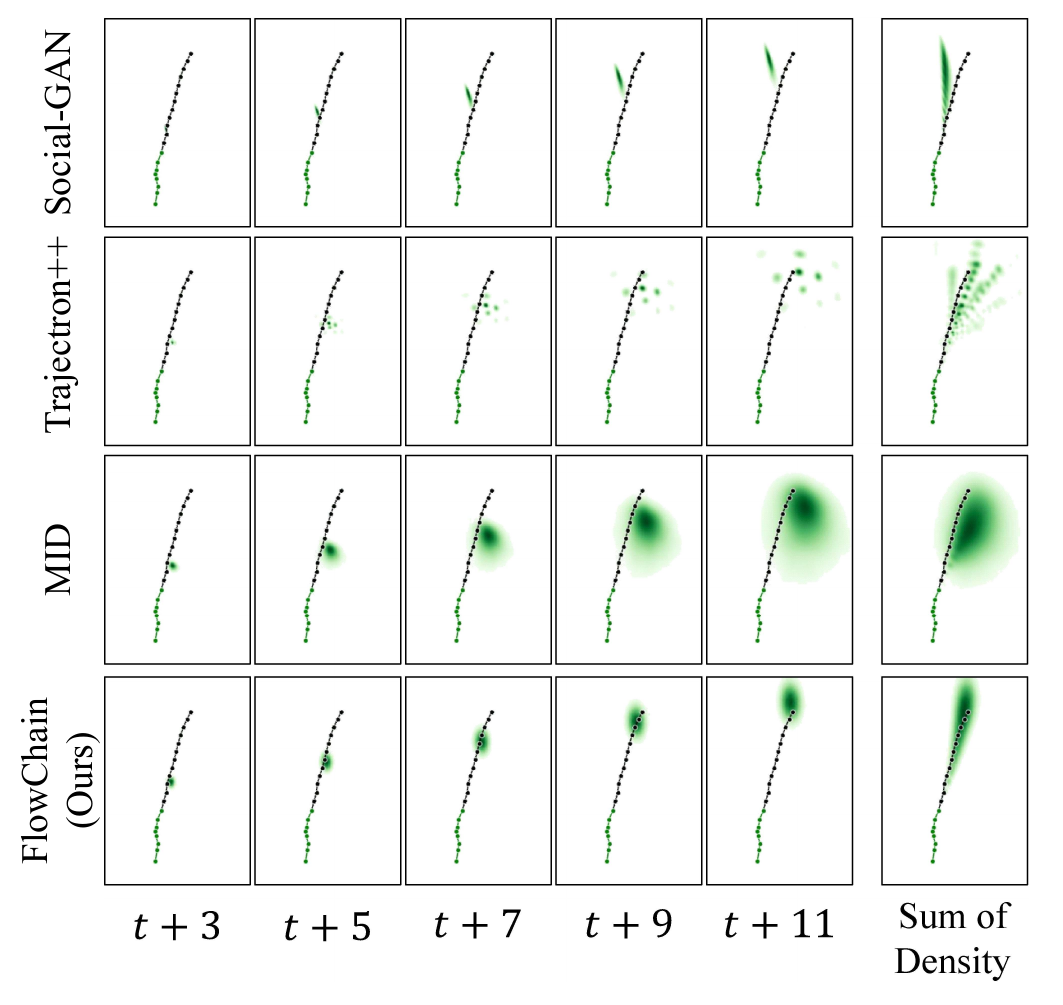}\\
    \vspace*{-2mm}
    \caption{
    \textbf{Temporal estimated densities on \textit{ETH/UCY} dataset.}
    }
    \label{fig:exp_eth}
\end{figure}

The results of density estimation are compared between FlowChain and previous methods based on VAE, GAN, and diffusion approaches.
Trajectron++ is the representative VAE-based trajectory prediction model with density estimation by GMM.
Social-GAN is a minimal model of GAN-based trajectory prediction.
MID is the diffusion-based trajectory prediction model with state-of-the-art prediction accuracy.
We applied KDE to Social-GAN and MID since they cannot estimate the density alone.
For a fair comparison, FlowChain, Social-GAN, and MID generated the same number of samples for future trajectories; \(10^5\) samples.

We evaluated the computational time and the accuracy of density estimation of each method, as shown in Tab.~\ref{tab:time}.
Social-GAN and MID take much longer time because KDE is slow in spite of running on GPU.
Trajectron++ is a bit faster than our FlowChain due to its simple Gaussian mixture model but suffers from the poor quality of estimated density, as shown by the worst EMD, 2.545, on \textit{Simfork} dataset and worst log-probability, \(-281.86\), on \textit{ETH/UCY} dataset.
On the other hand, our FlowChain achieved the best EMD, 1.408, and the best log-probability, -0.26, in spite of the fast computation \(37.0ms\).
Thus, our method successfully addresses both the computational cost and accurate density estimation.

The estimated densities on \textit{Simfork} are visualized in Fig.~\ref{fig:exp_simfork}.
The densities estimated by Trajectron++ are too large compared to the ground truth densities.
While the densities of Social-GAN and MID are stretched to the vertical axis, our FlowChain successfully estimated the tighter round-shape densities closest to the ground truth densities.

The estimated densities on \textit{ETH/UCY} are shown in Fig.~\ref{fig:exp_eth}.
%
The size of each density represents the positional uncertainty of a predicted position.
As within \textit{Simfork}, in \textit{ETH/UCY}, an agent cannot change its traveling direction significantly in a short period (i.e., $T_{f}=12$ is $0.4 \times 12$ = $4.8$ seconds), the size of the density expected to be tight.
The densities estimated by Social-GAN are tight but deviated from the ground truth trajectory.
Trajectron++ generated the densities with several disconnected modes due to the approximation as a Gaussian mixture.
While the densities from MID are close to the ground truth trajectory, MID could not suggest tight densities and their modes deviated from the ground truth trajectories.
On the contrary, the density from our FlowChain covers the ground truth trajectory with tight distributions.
thanks to the explicit density estimation of conditional CIFs and our expressive model architecture.
We should note that the estimated density of FlowChain has some fluctuation due to the looseness of the density computation of CIFs.
However, we did not observe any potential problem or failure of density estimation related to this in our experiments.

\begin{figure}[t]
    \centering
    \includegraphics[width=\linewidth]{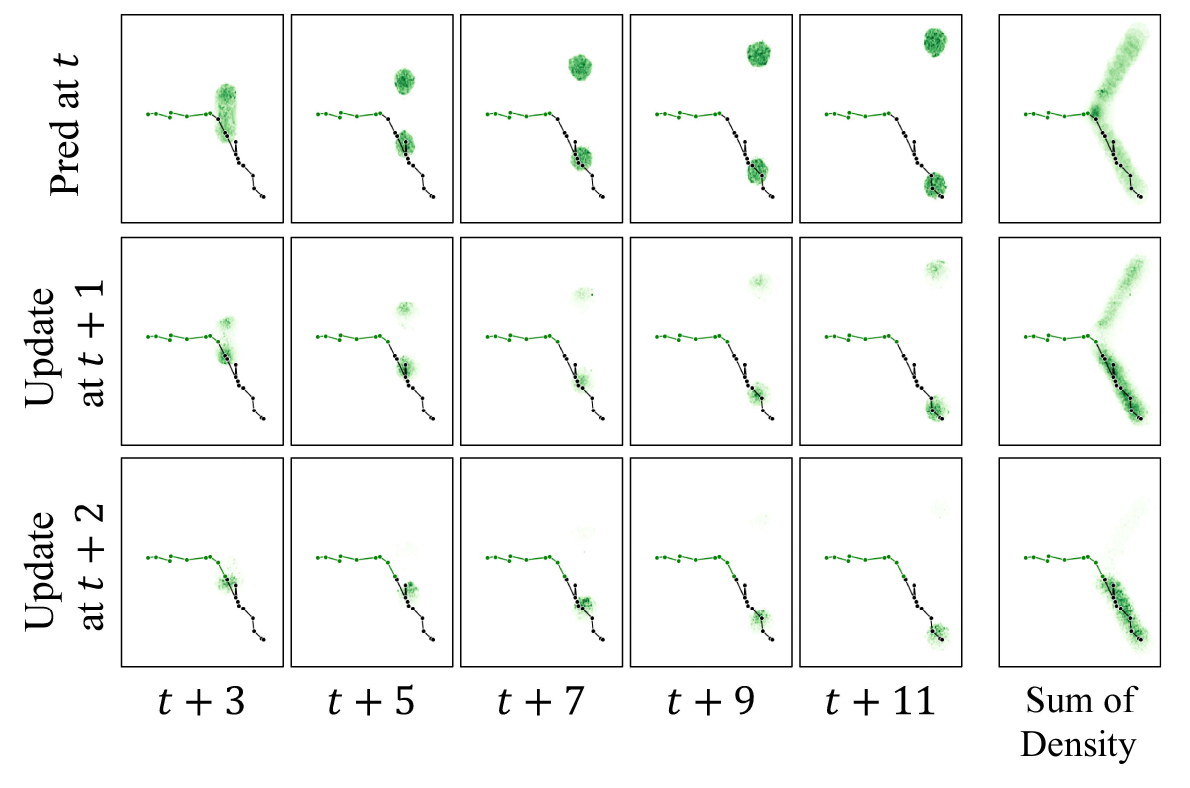}\\
    \vspace*{-2mm}
    \caption{
    \textbf{Temporal densities estimated by our update procedure on \textit{Simfork} dataset.}
    }
    \label{fig:exp_update_simfork}
    \vspace*{-2mm}
\end{figure}

\begin{figure}[t]
    \centering
    \includegraphics[width=\linewidth]{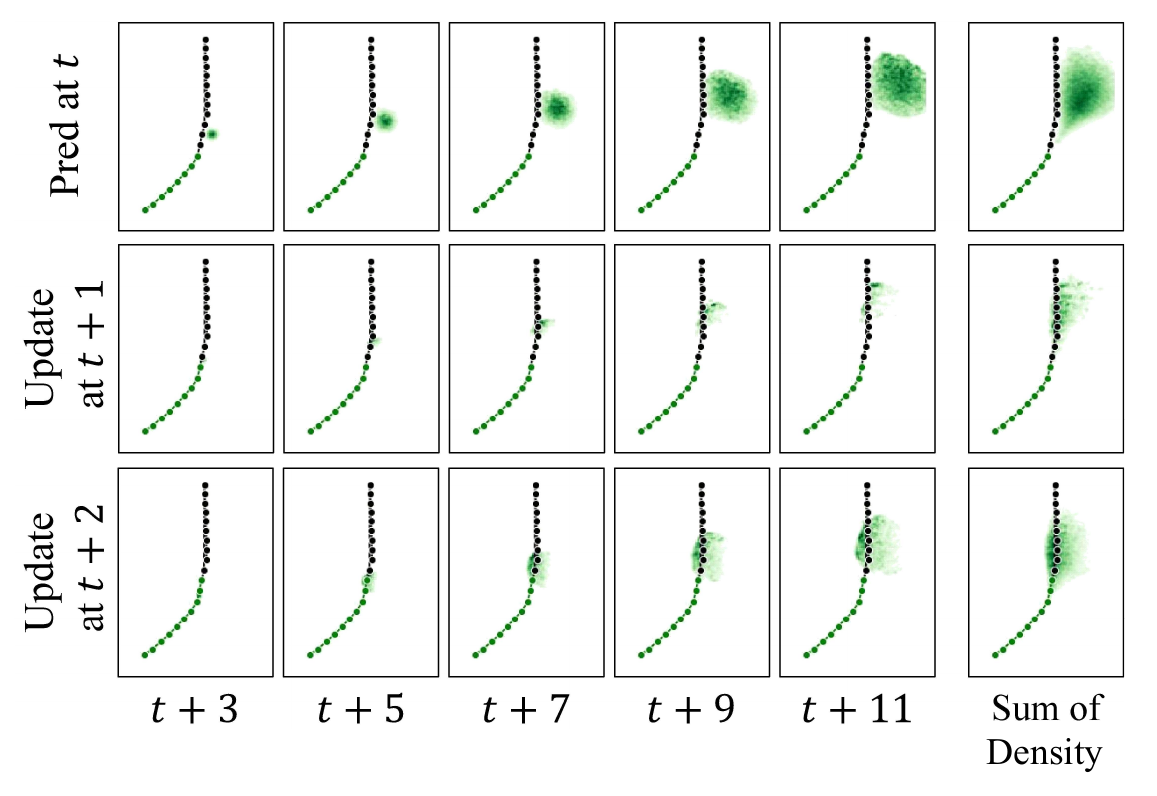}\\
    \vspace*{-2mm}
    \caption{\textbf{Temporal densities estimated by our update procedure on \textit{ETH/UCY} dataset.}}
    \label{fig:exp_update_eth}
    \vspace{-1mm}
\end{figure}

\begin{table*}[t]
    \centering
    \caption{\textbf{Accuracy and computational time of density estimation.}
    Scores are in milliseconds for computational time.
    While lower is better for EMD on \textit{Simfork} dataset, higher is better for log-probablity on \textit{ETH/UCY} dataset. 
    We report the averaged EMD and log-probability over the 12 prediction steps.
    }
\begin{tabular}{ll|rc|rr}
\hline
\multicolumn{2}{l|}{} & \multicolumn{2}{c|}{\textbf{Comp. Time}} & \multicolumn{2}{c}{\textbf{Accuracy of Density Estimation}} \\
\multicolumn{2}{l|}{\multirow{-2}{*}{\textbf{Method}}} & \multicolumn{1}{c}{\textbf{All}} & \textbf{w/o KDE} & \textbf{EMD $\downarrow$} & \textbf{log-probability $\uparrow$} \\ \hline
\multicolumn{2}{l|}{Social-GAN~\cite{gupta2018social}} & 11498.8 & \multicolumn{1}{r|}{9.8} & 1.544 & -3.40 \\
\multicolumn{2}{l|}{Trajectron++~\cite{salzmann2020trajectron++}} & 33.7 & - & 2.545 & -281.86 \\
\multicolumn{2}{l|}{MID~\cite{gu2022stochastic}} & 23904.1 & \multicolumn{1}{r|}{12219.1} & 2.135 & -0.90 \\
 & Pred at \(t\) & 37.0 & - & {\color[HTML]{FE0000} \textbf{1.408}} & {\color[HTML]{FE0000} \textbf{-0.26}} \\ \cline{2-6} 
 & Update from \(t-1\) & 0.6 & - & 2.539 & -0.45 \\
\multirow{-3}{*}{FlowChain (Ours)} & Update from \(t-5\) & 0.5 & - & 2.899 & -0.53 \\ \hline
\end{tabular}
    \label{tab:time}
    \vspace{-1mm}
\end{table*}

\begin{figure}
    \centering
    \includegraphics[width=0.9\linewidth]{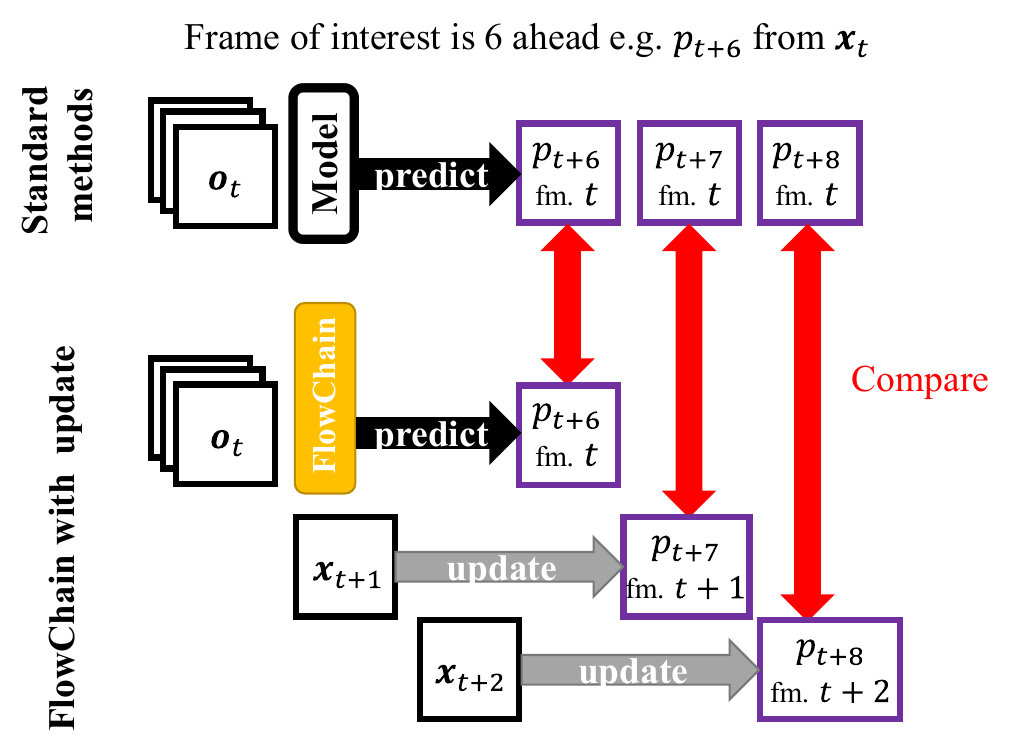}
    \caption{\textbf{Comparison of standard methods and FlowChain with updates.}}
    \label{fig:exp_update}
    \vspace{-2mm}
\end{figure}

\begin{table*}[]
    \centering
    \caption{\textbf{Accuracy of density estimation with the update.}}
\begin{tabular}{ll|rrrrrrrr}
\hline
\multicolumn{2}{l|}{\multirow{2}{*}{\textbf{Method}}} & \multicolumn{8}{c}{\textbf{log-probability on each step}} \\ \cline{3-10} 
\multicolumn{2}{l|}{} & \multicolumn{1}{c}{\textbf{t+6}} & \multicolumn{1}{c}{\textbf{t+7}} & \multicolumn{1}{c}{\textbf{t+8}} & \multicolumn{1}{c}{\textbf{t+9}} & \multicolumn{1}{c}{\textbf{t+10}} & \multicolumn{1}{c}{\textbf{t+11}} & \multicolumn{1}{c|}{\textbf{t+12}} & \multicolumn{1}{l}{\textbf{mean}} \\ \hline
\multicolumn{2}{l|}{Social-GAN~\cite{gupta2018social}} & -2.65 & -2.86 & -3.07 & -3.26 & -3.44 & -3.62 & \multicolumn{1}{r|}{-3.80} & -3.24 \\
\multicolumn{2}{l|}{Trajectron++~\cite{salzmann2020trajectron++}} & -121.19 & -93.61 & -71.62 & -51.36 & -36.85 & -27.92 & \multicolumn{1}{r|}{-22.72} & -97.79 \\
\multicolumn{2}{l|}{MID~\cite{gu2022stochastic}} & -1.06 & -1.43 & -1.75 & -2.02 & -2.26 & -2.46 & \multicolumn{1}{r|}{-2.65} & -1.95 \\
\multirow{2}{*}{FlowChain} & Without Update & 0.21 & 0.36 & \textbf{0.44} & -0.08 & -0.78 & -0.75 & \multicolumn{1}{r|}{-1.19} & -0.26 \\
 & With Update & \textbf{0.20} & \textbf{0.50} & 0.39 & \textbf{0.44} & \textbf{0.20} & \textbf{0.26} & \multicolumn{1}{r|}{\textbf{0.16}} & \textbf{0.31} \\ \hline
\end{tabular}
\label{tab:exp_update}
    \vspace{-1mm}
\end{table*}

\subsection{Proposed Update of Density Estimation}

As shown in Tab.~\ref{tab:time}, the accuracy of density estimation degraded along with the number of updates.
However, this update is remarkably fast \(0.6ms\), also shown in Tab.~\ref{tab:time}.
Thus, autonomous systems can select the accurate prediction from scratch or the super rapid update by reusing predictions depending on the utilization of their computational resources.

The estimated densities of our update procedure are shown in Fig.~\ref{fig:exp_update_simfork} and Fig.~\ref{fig:exp_update_eth}, which show the densities updated at \(t+1\) and \(t+2\) from the prediction at \(t\).
On \textit{Simfork} dataset, the wrong mode is successfully suppressed by exploiting the information of the new observation.
On \textit{ETH/UCY} dataset, the initial prediction deviated from the ground truth because the observed trajectory observed until \(t\) does not have the information on turning left.
However, the updated densities covered the ground truth well by adopting the \textit{newest observed position} despite the minimal computational cost.

We further perform a comparison with a practical setting.
Models usually make predictions with intervals due to the large computational cost.
Therefore, we usually cannot obtain the prediction from the latest observation, as shown by the upper side of Fig.~\ref{fig:exp_update}.
For example, assume we need the density of six time-step ahead such as \(p_{t+6}\) from \(\mathbf{x}_t\), we cannot obtain \(p_{t+7}\) from \(\mathbf{x}_{t+1}\) or \(p_{t+8}\) from \(\mathbf{x}_{t+2}\) due to the computational overhead.
However, with FlowChain's rapid update, we can obtain the update (\(p_{t+7}\), \(p_{t+8}\)) from the latest observation (\(\mathbf{x}_{t+1}\),\(\mathbf{x}_{t+2}\)) on every time step thanks to the minimal computational cost, as shown by the bottom side of Fig.~\ref{fig:exp_update}.
We conducted the comparison with the frame of interest as six time-step ahead \(t+6\) and the log-probability metric.
As shown by Tab.~\ref{tab:exp_update}, our FlowChain with update achieves the highest log-probability.
Therefore, we can conclude that FlowChain's update can improve the density estimation with minimal computational cost.

\section{Concluding Remarks}

In this paper, we proposed a new normalizing flow-based trajectory prediction network named FlowChain that achieves fast and accurate probability density estimation. Furthermore, our update procedure reliably generates the density estimation of the next step based on the density estimation of the previous step in an instant.
Experimental results demonstrated that FlowChain achieved comparable accuracy to the state-of-the-art trajectory prediction models and superiority in the accuracy of density estimation.

Our FlowChain has one limitation.
Our update procedure can reliably estimate the density for several steps but the performance will degrade along with the time steps because the reused flow transformations do not reflect the new observations.
Our future work is determining when should we stop to use the update procedure and make predictions from scratch.
Furthermore, FlowChain is the domain-agnostic model that can be applied to a wide range of prediction tasks such as stock price prediction, human motion prediction, and video prediction.
We will investigate the applicability of our FlowChain on these other domains.

{\small
\bibliographystyle{ieee_fullname}
\bibliography{egbib}
}

\clearpage

\section{Comparison with Flomo~\cite{scholler2021flomo}}

Although Flomo~\cite{scholler2021flomo} is the closest work to our FlowChain, we didn't include the results of Flomo in Sec.4, because the reproducible training code or pretrained model is unavailable.
Furthermore, Flomo reports the scores based on the train-validation-test splits different from ours.
Thus, we made a fair comparison along with the evaluation in the Social-GAN~\cite{gupta2018social} paper on \textit{ETH/UCY} dataset and Trajnet~\cite{sadeghian2018trajnet} paper on \textit{SDD} dataset.
We used the official implementation of Flomo for comparison.
Note that KDE is applied to the generated trajectories of Flomo for estimating temporal densities, as mentioned in Sec.~2 of the main paper.

The results of trajectory prediction accuracy are shown in Tab.~\ref{tab:supp_ethucy} and Tab.~\ref{tab:supp_sdd}.
Our FlowChain achieves better results than Flomo on most splits on \textit{ETH/UCY} and \textit{SDD}.

\begin{figure}[t]
    \centering
    \includegraphics[width=\linewidth]{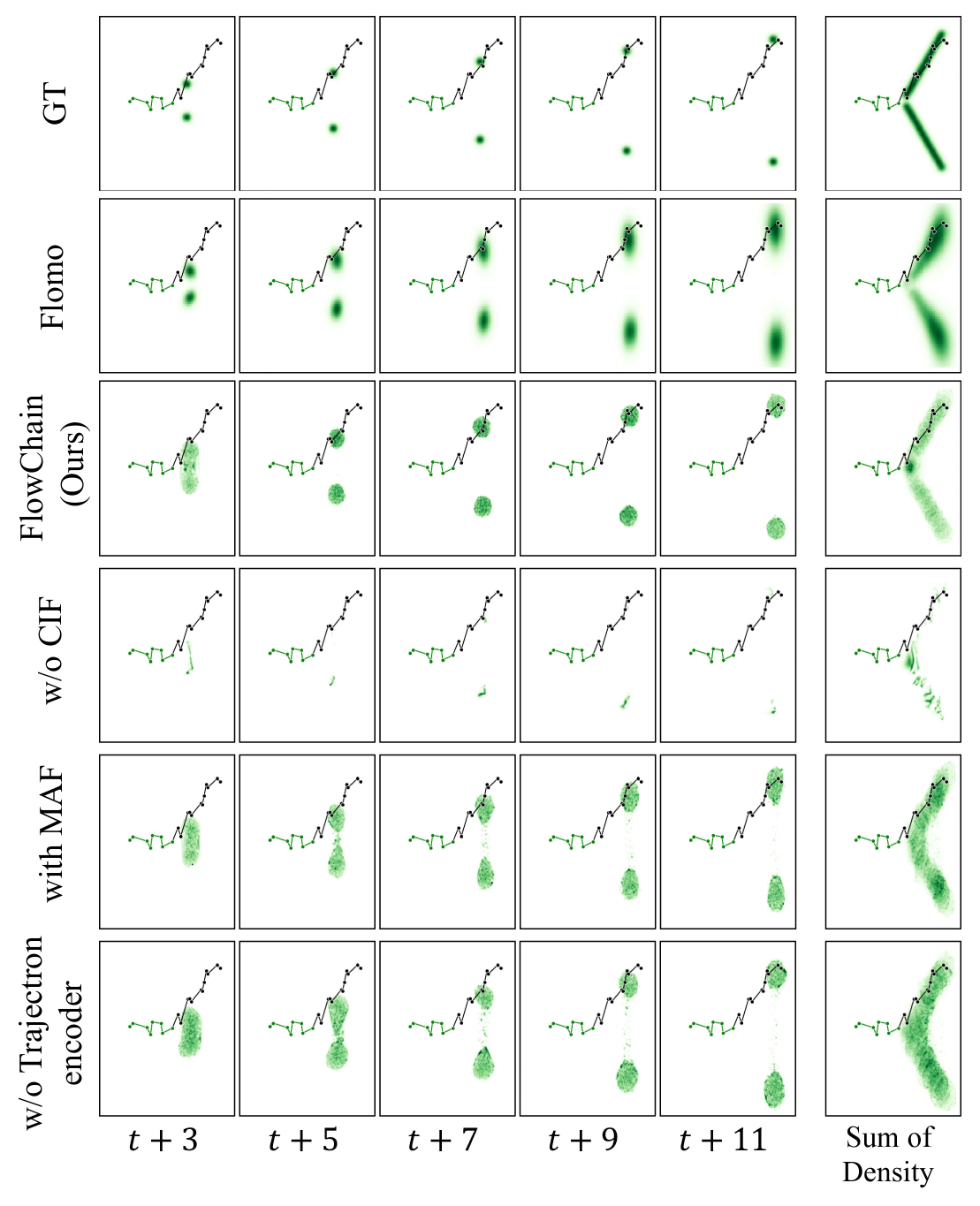}
    \caption{\textbf{Temporal estimated densities of Flomo and ablated models on \textit{Simfork} dataset.}}
    \label{fig:supp_simfork}
\end{figure}

\begin{figure}[t]
    \centering
    \includegraphics[width=\linewidth]{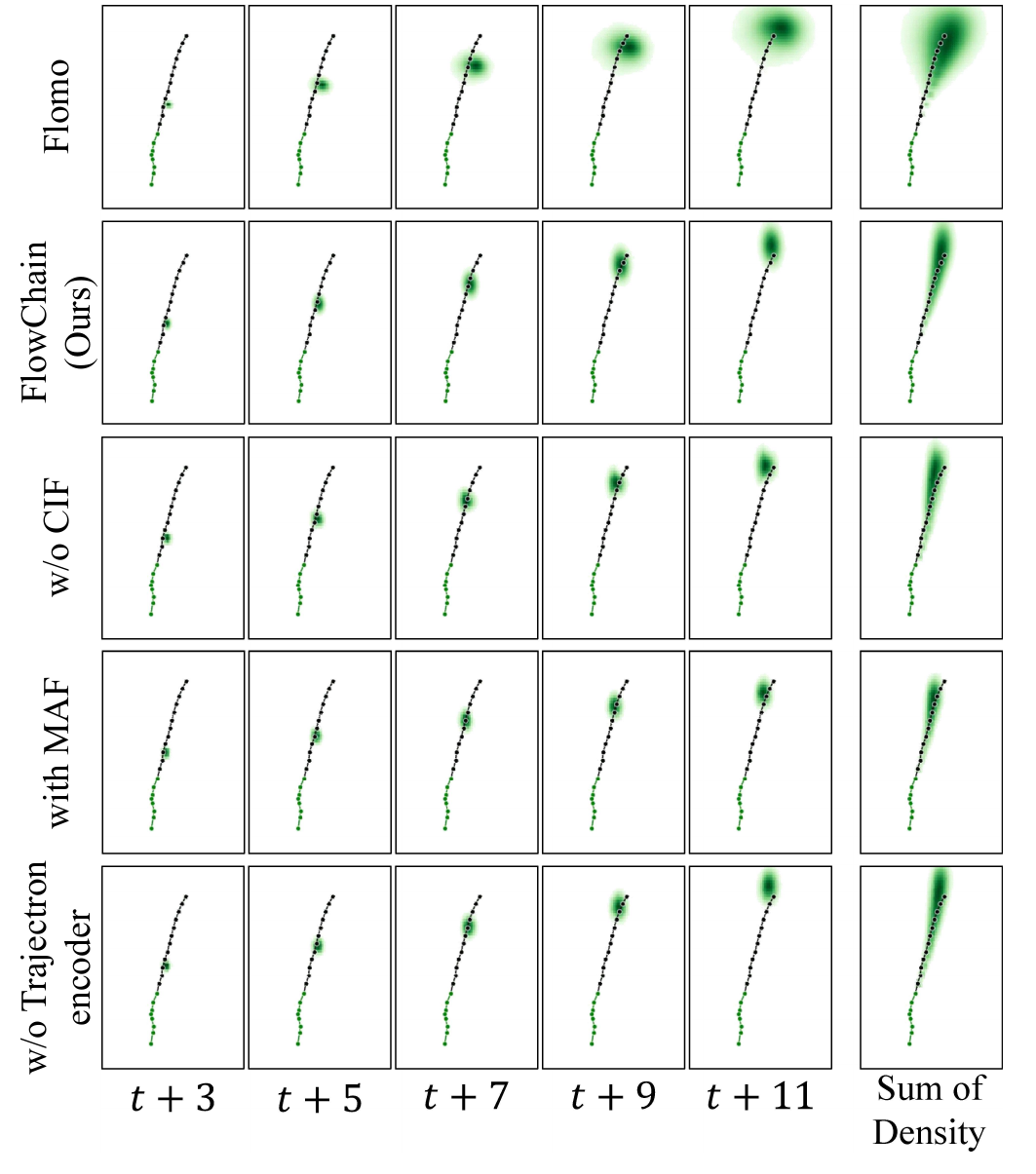}
    \caption{\textbf{Temporal estimated densities of Flomo and ablated models on \textit{ETH/UCY} dataset.}}
    \label{fig:supp_eth}
\end{figure}

The accuracy and computational time of density estimation of Flomo are compared with FlowChain as shown in Tab.~\ref{tab:supp_time}.
Flomo takes a much longer time for density estimation (i.e., 12397.4 milliseconds) because Flomo needs KDE for temporal density estimation.
Furthermore, the accuracy of density estimation of Flomo is lower than that of FlowChain over EMD on \textit{Simfork} dataset (i.e, 1,834 vs. 1.408) and log-probability on \textit{ETH/UCY} dataset (i.e, -24.05 vs. -0.26).
We also visualize the estimated densities of Flomo and FlowChain in Fig.~\ref{fig:supp_simfork} and Fig.~\ref{fig:supp_eth}.
On \textit{Simfork} dataset, Flomo estimated too large and elliptical-shape densities, unlike the tight round-shape ground truth distribution, as shown in the first row of Fig.~\ref{fig:supp_simfork}.
On \textit{ETH/UCY} dataset also, Flomo could not suggest tight densities.
These poor performances on density estimation are the limitations of KDE on limited samples.

\begin{table*}
    \centering
    \caption{\textbf{Quantitative comparison on \textit{ETH/UCY} dataset with Best-of-20 metrics.} Scores are in meters. Lower is better.}
    \begin{tabular}{l|rrrrrrrrrr|rr}
\hline
 & \multicolumn{2}{c}{\textbf{ETH}} & \multicolumn{2}{c}{\textbf{HOTEL}} & \multicolumn{2}{c}{\textbf{UNIV}} & \multicolumn{2}{c}{\textbf{ZARA1}} & \multicolumn{2}{c|}{\textbf{ZARA2}} & \multicolumn{2}{c}{\textbf{Mean}} \\ \cline{2-13} 
\multirow{-2}{*}{\textbf{Method}} & \multicolumn{1}{c}{ADE} & \multicolumn{1}{c}{\cellcolor[HTML]{EFEFEF}FDE} & \multicolumn{1}{c}{ADE} & \multicolumn{1}{c}{\cellcolor[HTML]{EFEFEF}FDE} & \multicolumn{1}{c}{ADE} & \multicolumn{1}{c}{\cellcolor[HTML]{EFEFEF}FDE} & \multicolumn{1}{c}{ADE} & \multicolumn{1}{c}{\cellcolor[HTML]{EFEFEF}FDE} & \multicolumn{1}{c}{ADE} & \multicolumn{1}{c|}{\cellcolor[HTML]{EFEFEF}FDE} & \multicolumn{1}{c}{ADE} & \multicolumn{1}{c}{\cellcolor[HTML]{EFEFEF}FDE} \\ \hline
Flomo~\cite{scholler2021flomo} & 0.58 & \cellcolor[HTML]{EFEFEF}1.02 & 0.34 & \cellcolor[HTML]{EFEFEF}0.63 & \textbf{0.29} & \cellcolor[HTML]{EFEFEF}\textbf{0.52} & 0.22 & \cellcolor[HTML]{EFEFEF}\textbf{0.39} & 0.28 & \cellcolor[HTML]{EFEFEF}0.53 & 0.34 & \cellcolor[HTML]{EFEFEF}0.62 \\
FlowChain (Ours) & \textbf{0.55} & \cellcolor[HTML]{EFEFEF}\textbf{0.99} & \textbf{0.20} & \cellcolor[HTML]{EFEFEF}\textbf{0.35} & \textbf{0.29} & \cellcolor[HTML]{EFEFEF}0.54 & \textbf{0.22} & \cellcolor[HTML]{EFEFEF}0.40 & \textbf{0.20} & \cellcolor[HTML]{EFEFEF}\textbf{0.34} & \textbf{0.29} & \cellcolor[HTML]{EFEFEF}\textbf{0.52} \\ \hline
FlowChain w/o CIF & 0.70 & \cellcolor[HTML]{EFEFEF}1.35 & 0.27 & \cellcolor[HTML]{EFEFEF}0.53 & 0.33 & \cellcolor[HTML]{EFEFEF}0.67 & 0.23 & \cellcolor[HTML]{EFEFEF}0.45 & \textbf{0.20} & \cellcolor[HTML]{EFEFEF}0.39 & 0.35 & \cellcolor[HTML]{EFEFEF}0.68 \\
FlowChain with MAF & 0.57 & \cellcolor[HTML]{EFEFEF}1.07 & 0.22 & \cellcolor[HTML]{EFEFEF}0.37 & \textbf{0.29} & \cellcolor[HTML]{EFEFEF}0.55 & 0.23 & \cellcolor[HTML]{EFEFEF}0.43 & 0.21 & \cellcolor[HTML]{EFEFEF}0.40 & 0.30 & \cellcolor[HTML]{EFEFEF}0.56 \\
FlowChain w/o Trajectron encoder & 0.62 & \cellcolor[HTML]{EFEFEF}1.23 & 0.34 & \cellcolor[HTML]{EFEFEF}0.72 & 0.33 & \cellcolor[HTML]{EFEFEF}0.62 & 0.23 & \cellcolor[HTML]{EFEFEF}0.42 & 0.21 & \cellcolor[HTML]{EFEFEF}0.39 & 0.35 & \cellcolor[HTML]{EFEFEF}0.68 \\ \hline
\end{tabular}
    \label{tab:supp_ethucy}
\end{table*}

\begin{table}
    \centering
    \caption{\textbf{Quantitative comparison on \textit{SDD} dataset with Best-of-20 metrics.} Scores are in pixels. Lower is better.}
    \begin{tabular}{l|r
>{\columncolor[HTML]{EFEFEF}}r }
\hline
\textbf{Method} & \multicolumn{1}{c}{ADE} & \multicolumn{1}{c}{\cellcolor[HTML]{EFEFEF}FDE} \\ \hline
Flomo~\cite{scholler2021flomo} & 10.78 & 17.36 \\
FlowChain (Ours) & 9.93 & 17.17 \\ \hline
FlowChain w/o CIF & \textbf{9.70} & \textbf{17.13} \\
FlowChain with MAF & 14.58 & 24.70 \\
FlowChain w/o Trajectron encoder & 18.92 & 30.94 \\ \hline
\end{tabular}
    \label{tab:supp_sdd}
\end{table}

\begin{table*}
    \centering
    \caption{\textbf{Accuracy and computational time comparison of density estimation.}
    Scores are in milliseconds for computational time.
    While lower is better for EMD on \textit{Simfork} dataset, higher is better for log-probablity on \textit{ETH/UCY} dataset. 
    We report the averaged EMD and log-probability over the 12 prediction steps.}
    \begin{tabular}{l|rc|rr}
\hline
\multirow{2}{*}{\textbf{Method}} & \multicolumn{2}{c|}{\textbf{Comp. Time}} & \multicolumn{2}{c}{\textbf{Accuracy of Density Estimation}} \\
 & \multicolumn{1}{c}{\textbf{All}} & \textbf{w/o KDE} & \textbf{EMD $\downarrow$} & \textbf{log-probability $\uparrow$} \\ \hline
Flomo~\cite{scholler2021flomo} & 12397.4 & \multicolumn{1}{r|}{24.2} & 1.834 & -24.05 \\
FlowChain (Ours) & 37.0 & - & \textbf{1.408} & \textbf{-0.26} \\ \hline
FlowChain w/o CIF & 26.7 & - & 3.313 & -9.48 \\
FlowChain with MAF & 44.2 & - & 2.475 & -0.55 \\
FlowChain w/o Trajectron encoder & 38.7 & - & 2.369 & -0.33 \\ \hline
\end{tabular}
    \label{tab:supp_time}
\end{table*}

\section{Effectiveness of the FlowChain Architecture}

We validated the effectiveness of the FlowChain architecture by changing each component.
We replace our three choices: CIF~\cite{cornish2020relaxing}, RealNVP~\cite{dinh2017density}, and Trajectron~\cite{salzmann2020trajectron++} encoder.
As mentioned in Sec.~3 of the main paper, we use conditional CIFs as the base normalizing flow model for better expressive power.
For replacing CIF, vanilla conditional normalizing flow models are employed, denoted by ``w/o CIF''.
As mentioned in Sec.~4 of the main paper, we use a three-layer RealNVP inside the CIF and Trajectron++ encoder for a temporal-social encoder.
For replacing RealNVP choice, we choose the MAF~\cite{papamakarios2017masked} instead of RealNVP, denoted by ``with MAF''.
For replacing the Trajectron++ encoder, one transformer layer is applied for encoding the observed past trajectory, denoted by ``w/o Trajectron encoder''.
Note that this transformer layer encodes each trajectory independently and thus cannot account for the social interactions.

The results of trajectory prediction accuracy are shown in Tab.~\ref{tab:supp_ethucy} and Tab.~\ref{tab:supp_sdd}.
By replacing CIF, the accuracy degrades on \textit{ETH/UCY} but gets slightly better on \textit{SDD}.
This is because the multi-modal densities are less observed in \textit{SDD} and thus models without CIF still fit the \textit{SDD} dataset.
However, as shown in Fig.~\ref{fig:supp_simfork}, models without CIF do not fit the multi-modal density and estimate thin densities.
We further confirmed the poor accuracy of density estimation without CIF by comparing the accuracy of density estimation, as shown in Tab.~\ref{tab:supp_time}.
We choose conditional CIFs as the base normalizing flow models because our main focus is on better and faster density estimation.
As mentioned in Sec.~3.2 of the main paper, we can see a little fluctuation in the estimated densities by models with CIFs.
However, this fluctuation can be ignored because estimated densities are smooth enough especially for the real data like \textit{ETH/UCY} dataset, as shown in Fig.~\ref{fig:supp_eth}.

By replacing RealNVP, the accuracy degrades on both \textit{ETH/UCY} and \textit{SDD}.
We find RealNVP a good fit for our chaining architecture.

By replacing the Trajectron encoder, the accuracy degrades especially on \textit{SDD} because the simple transformer encoder cannot consider the social interactions as mentioned above.
Furthermore, the Trajctron encoder needs less computational cost than a simple transformer encoder despite accounting for the social interactions, as shown in Tab.~\ref{tab:supp_time}.

\section{Experiment on Unicycle Motion Dataset.}
\begin{figure*}[t]
    \centering
    \includegraphics[width=0.8\textwidth]{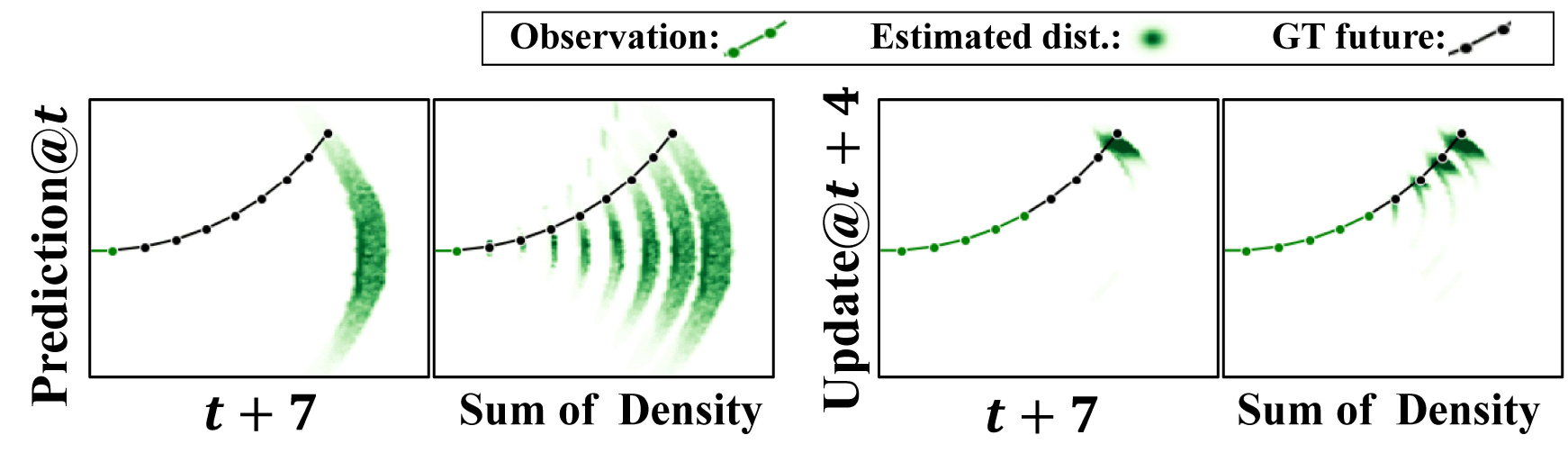}
    \caption{\textbf{Temporal estimated densities on the synthesized unicycle motion dataset.}}
    \label{fig:unicycle}
\end{figure*}

Our Simfork dataset, which only has simple bifurcations, cannot fully show the advantage of our method over Gaussian Mixture Models.
Therefore, we created a synthesized dataset based on the unicycle motion model. 
As shown in Fig~\ref{fig:unicycle}, FlowChain successfully captures the challenging banana-like shape distributions and produces reliable updates, which is hard for a mixture Kalman filter assuming Gaussian distributions.

\end{document}